\documentclass[conference]{IEEEtran}

\usepackage{times}
\usepackage{latexsym}
\usepackage{graphicx}
\usepackage{adjustbox}
\usepackage{float}
\usepackage{hyperref}
\usepackage{soul,color}
\usepackage[justification=centering]{caption}

\usepackage{fancyhdr}
\usepackage{lastpage}

\usepackage{amssymb}
\usepackage{hyperref}
\usepackage{color,soul}

\begin{document}




\title{Comprehending Linguistic and Emotional Cues in Demographically Diverse Spatial Social Media Data}

\title{Comprehending Lexical and Affective Ontologies in the Demographically Diverse Spatial Social Media Discourse}

\author{\IEEEauthorblockN{ Salim Sazzed} \\
\IEEEauthorblockA{\textit{Department of Computer Science,}
\textit{Old Dominion University,}
Norfolk, VA USA}

\IEEEauthorblockA{\textit{Department of Computer Science,}
\textit{University of Memphis,}
Memphis, TN, USA \\ saim.sazzed@gmail.com}
}

\maketitle

\begin{abstract}

This study aims to comprehend linguistic and socio-demographic features, encompassing English language styles, conveyed sentiments, and lexical diversity within spatial online social media review data. To this end, we undertake a case study that scrutinizes reviews composed by two distinct and demographically diverse groups. Our analysis entails the extraction and examination of various statistical, grammatical, and sentimental features from these two groups. Subsequently, we leverage these features with machine learning (ML) classifiers to discern their potential in effectively differentiating between the groups. Our investigation unveils substantial disparities in certain linguistic attributes between the two groups. When integrated into ML classifiers, these attributes exhibit a marked efficacy in distinguishing the groups, yielding a macro F1 score of approximately 0.85. Furthermore, we conduct a comparative evaluation of these linguistic features with word n-gram-based lexical features in discerning demographically diverse review data. As expected, the n-gram lexical features, coupled with fine-tuned transformer-based models, show superior performance, attaining accuracies surpassing 95\% and macro F1 scores exceeding 0.96. Our meticulous analysis and comprehensive evaluations substantiate the efficacy of linguistic and sentimental features in effectively discerning demographically diverse review data. The findings of this study provide valuable guidelines for future research endeavors concerning the analysis of demographic patterns in textual content across various social media platforms.
\end{abstract}






\section{Introduction}
Demographic data concerning user traits enable an understanding of user behaviors, ultimately facilitating improved decision-making for various social and business challenges. For instance, the analysis of demographic data assists businesses in making informed decisions regarding marketing, product development, customer experiences, and competitive positioning. In a social context, diverse demographic review data empowers policymakers to gain insights into the experiences and perspectives of different social groups, which, in turn, facilitates decision-making that promotes equity, inclusive representation, targeted interventions, evidence-based decision-making, and accountability. Furthermore, demographically tagged social media plays a significant role in computational social science by highlighting differences in beliefs and behaviors among demographic groups \cite{wood2017does}.

Recent studies have shown that demographic traits can be inferred from the linguistic characteristics of written content \cite{guimaraes2017age, chen2015comparative, hsieh2018author, dias2020cross}. The determination of various demographic attributes of users, such as age or gender, from written comments, has been investigated by \cite{guimaraes2017age, nguyen2014gender}. For example, Rosenthal and McKeown \cite{rosenthal2011age} attempted to predict the users' ages from blog content by incorporating various features specific to the blog and the behavior and interest of users. Schler et al. \cite{schler2006effects} observed significant differences in content and style levels between male and female bloggers by analyzing a large corpus of blogs of around 300 million words.

In addition, some studies tried to identify the English language nativeness of the writers from demographically diverse data. Although the perspective of their study was the second language acquisition (SLA) research, such as contrastive analysis, syntactic or grammatical errors made by non-native speakers \cite{wong2009contrastive,koppel2005automatically} based on corpus compiled from the sample essay of ESL (English as a Second Language) learners such as TOEFL (Test of English as a foreign
language) \cite{blanchard2013toefl11}, the international corpus of learner English \cite{granger2003international}. Research pertaining to demographically diverse informal reviews, such as those found in social media and prominently affected by English language nativeness and fluency level, remains mostly unexplored \cite{sarkar2020non} (except a few \cite{sazzed2022impact, sazzed2021hybrid, sazzed2022stylometric}).

Therefore, in this study, we aim to understand how linguistic and semantic attributes of text vary across demographically different groups in the context of social media, taking into account factors such as English language nativeness, geography, and socio-culture. In particular, we aim to provide insight into the following research questions-

\begin{itemize}    
    \item RQ1: Do the variations of the linguistic features (e.g., synthetic, lexical) render sufficient signals to distinguish diverse demographic groups when incorporated in classical ML classifiers?

    \item RQ2:  Whether the linguistic or n-gram lexical features perform better for the demography prediction task when incorporated into ML classifiers.
    
    

\end{itemize}

The two distinct demographic groups considered here represent individuals of two different socio-economic cultures. More importantly, the English reviews written by these two groups differ significantly in terms of English language nativeness and fluency levels. The first review group represents restaurant review data collected from Bangladesh, a country with mostly low proficient non-native English speakers \footnote{\url{https://www.ef.com/wwen/epi/}}. In Bangladesh, almost 98\% of people are Bengali native speakers \footnote{\url{https://www.worldatlas.com/articles/what-languages-are-spoken-in-bangladesh.html}}. The other group contains restaurant reviews written by users located in the USA (mostly English native speakers). We extract various statistical and syntactic features such as review length, frequency of opinion words, and usage of POS (part-of-speech) from the reviews of both groups. We find that linguistic features exhibit sufficient distinguishing signals to differentiate between the two groups of reviews when used as input for classical machine learning (CML) classifiers. Additionally, we explore the performance of lexical word n-grams-based features for classifying review groups by incorporating them into classical ML classifiers and transformer-based language models. As expected, we observe that transformer-based models, when utilizing lexical features, outperform both types of feature-based classical ML classifiers.





\subsection{Contributions}
The main contributions of this study can be summarized as follows-
\begin{enumerate}
 \item  We identify differences in stylistic, syntactic, and statistical features among reviews from diverse demographic groups.
 
 \item  We show that it is possible to distinguish demographically diverse informal social media reviews by incorporating various linguistic features into the machine learning classifiers.

 \item  Finally, we compare the differences in the performance of linguistic and lexical n-gram-based features for distinguishing diverse review groups.

\end{enumerate}

\begin{table}[ht]
\caption{Sample reviews from \textit{Demography-1} and \textit{Demography-2}}
\centering
\begin{tabular}{c}
\includegraphics[scale=0.54]{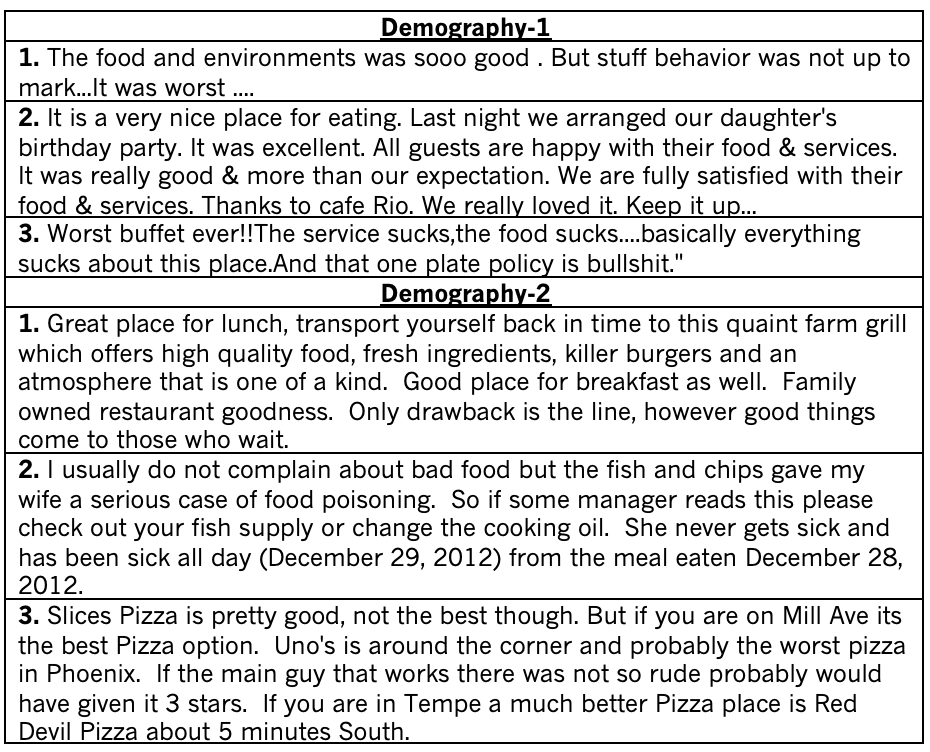} \\
\end{tabular}
\label{tab:gt}
\end{table}

\begin{figure}
\centering
\includegraphics[width=1.0 \linewidth]{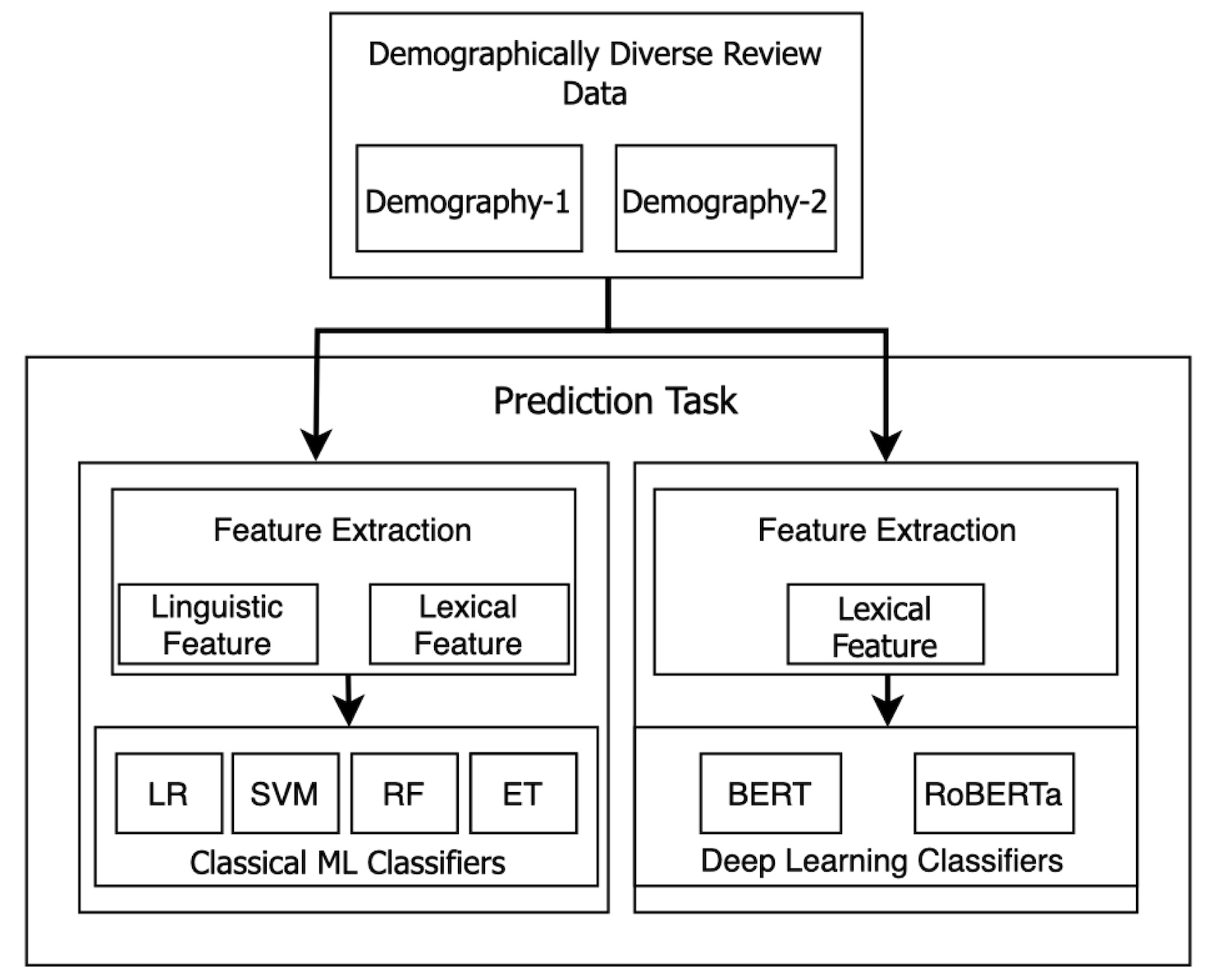}
\caption{Overall framework of the study}
\end{figure}

\section{Demography Prediction Task}

\subsection{Dataset}
As mentioned earlier, in this study, we investigate restaurant reviews written in English from two different demographic groups. The first group, which we refer to as \textit{Demography-1}, consists of reviews collected from restaurants located in Bangladesh \footnote{\url{https://www.kaggle.com/tuxboy/restaurant-reviews-in-dhaka-bangladesh}}. In Bangladesh, the majority of people (98\%) speak standard Bengali or one of its many dialects as their first language, while the remaining population speaks regional or minority languages. English proficiency in Bangladesh is generally categorized as low among secondary speakers \footnote{\url{https://www.ef.com/wwen/epi/}}. The second review group, referred to as \textit{Demography-2}, comprises a subset of the Yelp restaurant review dataset written by primarily English native speakers residing in the USA.



To mitigate potential domain bias in the evaluation process, the textual content used in both groups was sourced exclusively from the same domain, namely, restaurant reviews. Additionally, an equal number of reviews were selected for each group to prevent the influence of class imbalance on the classification outcomes. The final dataset comprises a total of 9974 reviews, where each group includes 4987 reviews.


\subsection{Classical Machine Learning-based Prediction}

To distinguish the two demographic groups, we employ several classical ML classifiers: Logistic Regression (LR),  Support Vector Machine (SVM), Random Forest (RF), and Extra Trees (ET). For all classical ML classifiers, the default parameter settings of the scikit-learn library \cite{scikit-learn} are used.

As an input of classical ML classifiers, we utilize two distinct groups of features: i) linguistic features and ii) word unigram and bigram-based lexical features, separately. The linguistic features further encompass several sub-types, including grammatical and sentiment features.   
 

\subsubsection{Linguistic Features}

We extract three different types of linguistic features from the reviews, which are utilized as input for the classical ML classifiers.
Text statistics such as review length in terms of words and sentences have been employed in related work such as language variety identification task \cite{van2017exploring}. Grammatical features such as the usage of PoS tags have been studied for language nativeness identification tasks in the earlier works as these stylistic features reflect user communication behavior and interaction style \cite{volkova2017identifying}. In addition, lexicon coverage, a feature that refers to the usage of opinion or sentiment words, is considered.

We employ the Mann-Whitney U test to determine which linguistic features show significant differences in the two groups of reviews. The Mann-Whitney U test is a non-parametric test of the null hypothesis, which is often used to test the differences in the distributions of two sets of values. We utilize the Mann-Whitney U test between two groups for all the linguistic features. The null hypothesis states that the distribution of a specific attribute in \textit{Demography-1}  is the same as the underlying distribution of the same attribute in \textit{Demography-2}, while the alternative hypothesis suggests the opposite.


\begin{table}[!ht]

\caption{The statistics of various linguistic features in the reviews belong to two groups,  \textit{Demography-1} and \textit{Demography-2}}
\centering
\resizebox{0.5\textwidth}{!}{\begin{tabular}{c|ccc}
\textbf{Type} &
\textbf{Feature} & \textbf{Demo-1}	& \textbf{Demo-2}	\\
\hline
 & Total words (in corpus) &  147401 &  656672	
\\
Text & Total sentences (in corpus) &  16272	&  47922	\\

\cline{2-4}
Statistics & Mean review length 
  (\#words) & 
 29.56	& 131.70
\\
\cline{2-4}
 & Mean review length (\#sentences)  & 3.32	& 9.611\\

\cline{2-4}
 & Mean sentence length (\#words) 
 & 8.89 &  13.70
\\
\hline
\hline

Lexical  & Total unique words (in corpus) & 10473 &  28297
\\
 Diversity &  &  &  
\\
\hline
 & Total negation words (in corpus) & 1585 &  5287
\\
Negation & (\%) of negation words ( in corpus ) & 1.0\% & 0.80\%
\\
 & Negation words per review (mean) & 0.32 & 1.06

\\
\hline
 & Total articles (in corpus) & 6553 &  45918
\\
 & (\%) of article in corpus (word) & 4.44\% & 6.99\%
 \\& Articles/review (words) & 1.31 &  9.20
\\

\cline{2-4}
Grammatical & Total adjectives (in corpus) & 15632 & 57561
\\
Feature &  (\%) of the adjective (in corpus) & 10.60\%   & 8.76\%
\\
&Number of adjectives/review & 3.13 & 11.54 \\

\cline{2-4}
 & Total verbs (in corpus)  & 14754 & 74881
\\
 & (\%) of the verb in corpus  & 10.01\%  & 11.40\%
\\
 & Number of verbs/review  & 2.95 & 15.01
\\
\cline{2-4}
 & Total prepositions (in corpus)  & 8466 & 52637
\\
 &  (\%) of the preposition (in corpus) & 5.79\% & 8.00\%
\\
 & Number of prepositions/review  & 1.69 & 10.55
\\

\cline{2-4}
 & Total SC(in corpus) & 2960 &  19402 \\
  & (\%)of the SC(in corpus) & 2.0\% & 2.95\%  \\
 & SC per reviews & 0.59 &  3.89 \\

\hline
\hline
 & Total opinion words (Hu \&Liu) & 8799 & 33691 \\
 Sentiment & Lexicon coverage (Hu \& Liu) & 6.22\% & 5.14\% \\
 Lexicon & Total opinion words (VADER) & 9516 & 36390 \\
 & Lexicon coverage (VADER)  & 6.51\% & 5.56\% 
\end{tabular}}
\label{content-independent-statistics}
\end{table}

\paragraph{ Text Length Features}
These features represent the length of the reviews based on word and sentence level: i) Number of words per review, ii) Number of sentences per review, iii) Number of words per sentence. As earlier studies suggested that English texts written by non-native English speakers are usually simpler than those of natives \cite{goldin2018native}, the sentence length could be a distinguishing factor for the demography prediction task.

    
    

\paragraph{Grammatical and Negation Features}

We consider a set of grammatical attributes that may provide signals to discern the diverse demographic groups.
\begin{itemize}

     \item Articles: The number of articles  (i.e., \textit{a, an, the}) present in a review is computed. 
     
     \item Adjectives:  The number of adjectives present in a review is computed. The spaCy \cite{spacy2} library is used to identify adjectives in a text.
     
     \item Verbs: The usage of verbs in both corpora is provided. Similar to adjective identification, spaCy \cite{spacy2} library is used for verb identification.

     \item Prepositions: We calculate the number of prepositions present in the reviews of two groups.  A list of commonly used prepositions is considered (details can be found here \footnote{\url{https://github.com/sazzadcsedu/LinguisticAnalysis}}).

    
    \item Subordinating conjunctions (SC): Additionally, we take into account the presence of subordinating conjunctions that indicate complex sentences. A complex sentence typically consists of one or more dependent (subordinate) clauses and one or more independent clauses. Subordinating conjunctions are words or phrases that connect dependent clauses to independent clauses. Examples of subordinating conjunctions include "although," "as," "because," "before," "how," "if," "once," "since," and so on. We examined the occurrence of 50 commonly used subordinating conjunctions in each review \footnote{\url{https://github.com/sazzadcsedu/LinguisticAnalysis}}.

    \item Negative words: The VADER \cite{Hutto2014VADERAP} negative word list is used as a reference to find the number of negative words in each group \footnote{\url{https://github.com/cjhutto/vaderSentiment} \label{githublink}}.

\end{itemize}

\paragraph{Sentiment Lexicon Coverage}

The coverage of two popular English sentiment lexicons, Opinion Lexicon \cite{Hu:2004:MSC:1014052.1014073} and VADER \cite{Hutto2014VADERAP}, is computed for each of the reviews from both groups.

    
    
    
    
    
    




\subsubsection{Lexical n-gram Features}

In addition, as lexical features, we extract word n-grams from the reviews of both groups. An n-gram denotes a consecutive sequence of n items within a given textual context. Particularly, we extract both unigrams (individual words) and bigrams (two-word combinations) from the review texts. Following this extraction, we calculate their respective term frequency-inverse document frequency (tf-idf) scores and incorporate these computed scores as input features into the CML classifiers.

\begin{figure*}
\centering
\includegraphics[width=0.99 \linewidth]{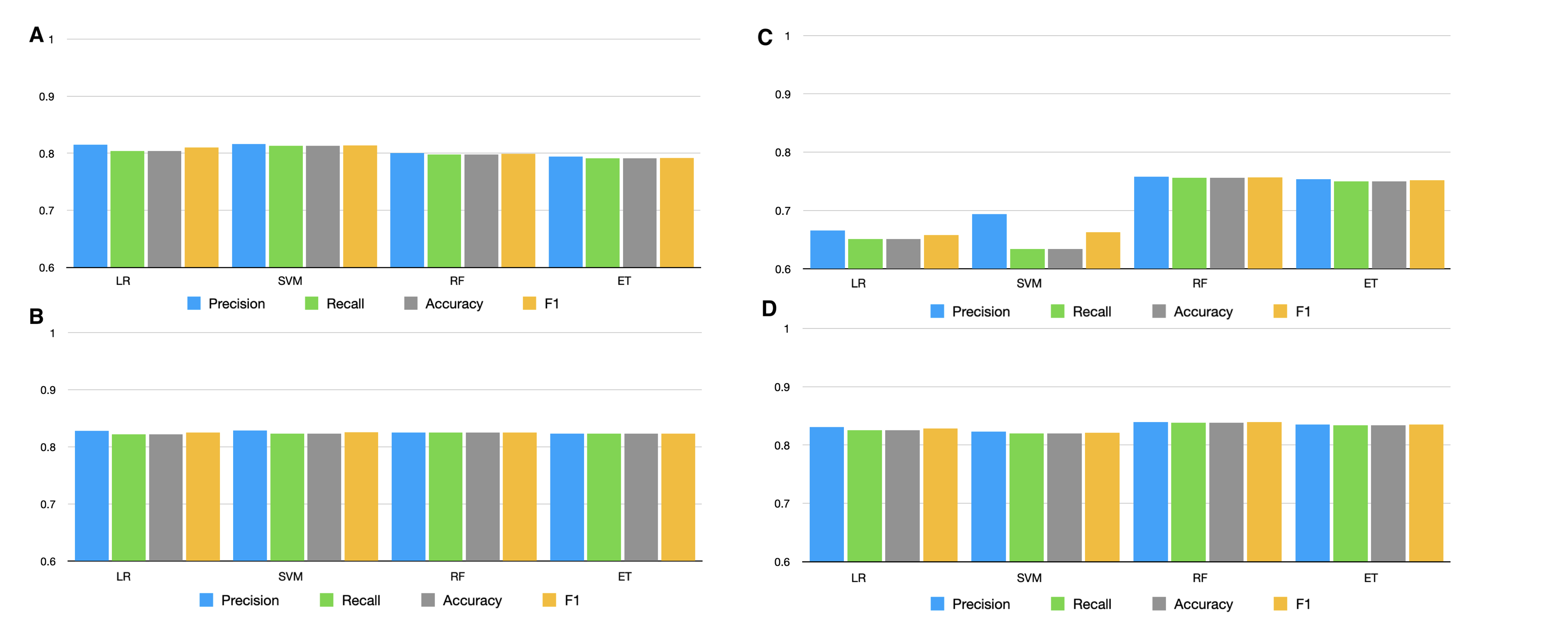}
\caption{Performances of classical ML classifiers using various types of linguistic features, (A) text statistics features, (B) grammatical features, (C) sentiment features, and (D) using features A, B, and C altogether}
\label{linguisti_classifiers}
\end{figure*}

\begin{figure}
\centering
\includegraphics[width=0.99 \linewidth]{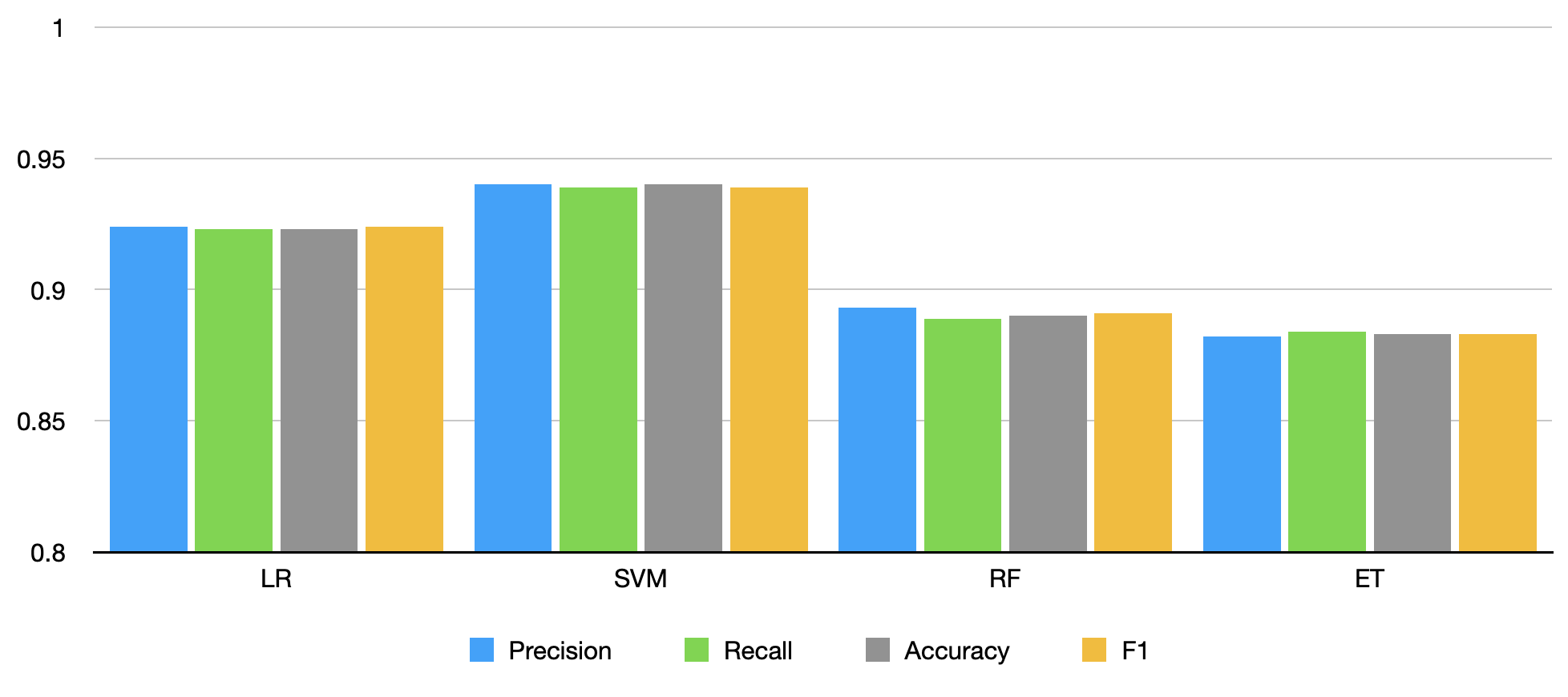}
\caption{Performances of classical ML classifiers using lexical features (i.e., word unigrams and bigrams)}
\label{lexical_cml}
\end{figure}

\subsection{Deep Learning-based Classification}
The transformer-based pre-trained language models, such as BERT \cite{devlin2018bert} and RoBERTa \cite{liu2019roberta}, have shown state-of-the-art results in various text classification tasks with limited labeled data. We fine-tune both transformer-based pre-trained models for categorizing reviews into two demographically diverse groups. Since this is a binary-level classification task, we utilize the classification module of these pre-trained models. The Hugging face library \cite{wolf-etal-2020-transformers} is used for fine-tuning all the pre-trained models. As the initial layers of pre-trained models primarily learn general features, they are left unchanged during fine-tuning process. Only the last layer of the pre-trained model is trained using new data specifically for the binary-level classification task.

We tokenize the input data for fine-tuning the language model. As pre-trained models typically support texts with a maximum of 512 tokens, we divide reviews longer than 512 words into 512-word chunks. During training, all the 512-token chunks are assigned the same class as the original review. During testing, the final class of the review is determined by majority voting. In the event of a tie, we consider the word length of the chunks to decide the final class label. For training, we use a mini-batch size of 8 and a learning rate of 4*10\textsuperscript {-5}. During the training process, 20\% of the samples are dedicated to validation. We optimize the pre-trained models using the Adam optimizer, with the loss parameter set to categorical cross-entropy. The training procedure is carried out for four epochs with an early stopping criterion set.  Note that all hyperparameters of deep learning models are determined based on empirical evaluation.











\begin{figure}
\centering
\includegraphics[width=0.99 \linewidth]{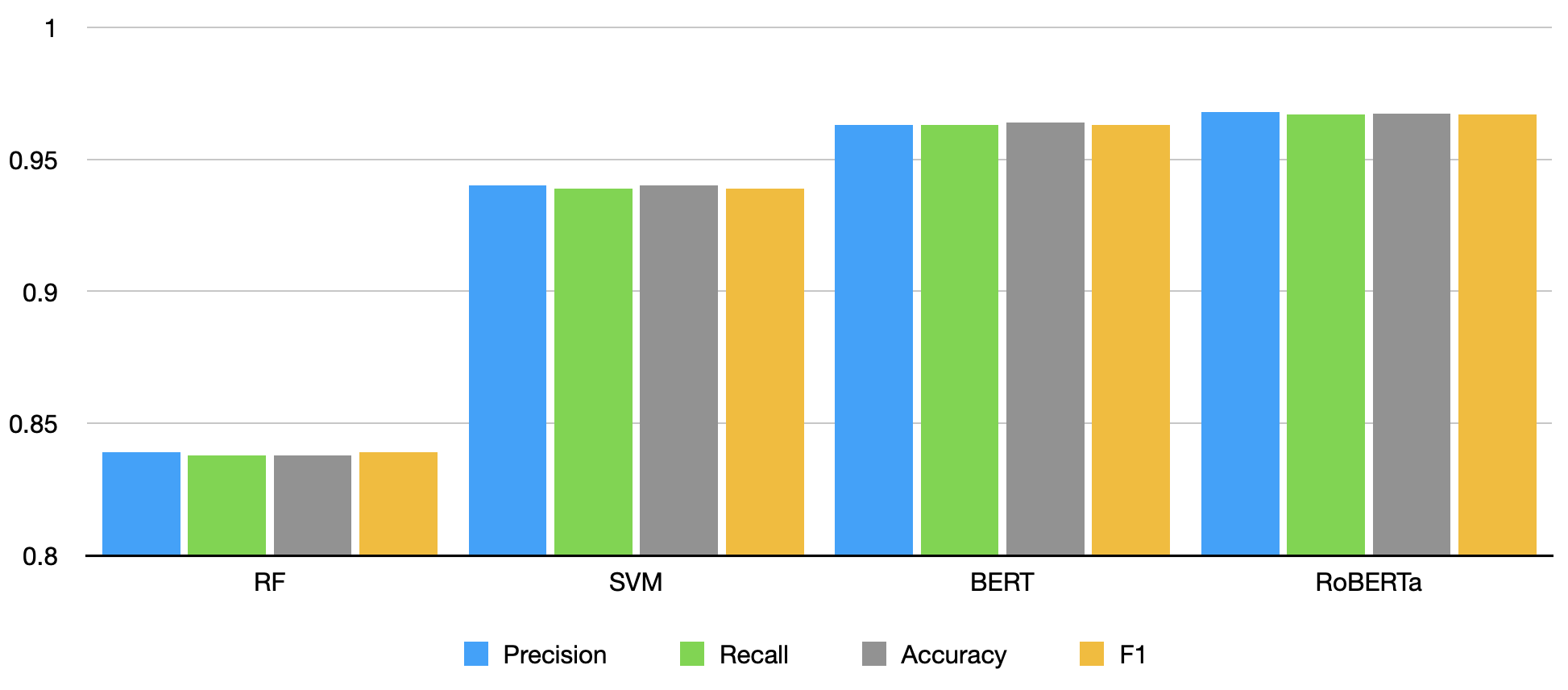}
\caption{The comparisons between the best results achieved by classical ML classifiers (i.e., RF with linguistic features and SVM with lexical features), and two transformer-based language models, for the demographic group prediction task}
\label{lexical_all}
\end{figure}


\section{Results and Discussion}

\subsection{Evaluation Settings}
To compare different classification methods for the demography prediction task, we utilize 5-fold cross-validation. Several key metrics, namely precision, recall, macro F1 score, and accuracy, are employed to assess the performance of classifiers. These metrics provide a comprehensive assessment of the performance of the different classifiers.




\begin{table}[!ht]

\caption{Top adjectives(\#occurrences) and verbs (\#occurrences)}
\centering

\resizebox{0.48\textwidth}{!}{
\begin{tabular}{cccc|cccc}
\\
 \multicolumn{4} {c}{\textbf{Top adjectives}} &\multicolumn{4} {c}{\textbf{Top verbs}}  
\\
\hline
 \multicolumn{2} {c}{Demography-1}   &   \multicolumn{2} {c}{Demography-2}  &  \multicolumn{2} {c}{Demography-1} &  \multicolumn{2} {c}{Demography-2} \\
\hline

good & 1609 & good & 2854 &  had & 1609 & had & 2987 \\
\hline
bad & 407 & great & 1925 & have & 363 & have & 2489 \\
\hline
worst & 331 & other & 1073 & go & 361 &  get & 1818 \\
\hline

best & 327 & little & 944 & ordered & 272 & go & 1609 \\
\hline
great & 274 & nice & 884 &serve & 242 & got & 888 \\
\hline
worst & 264 & more & 822 & went & 218 & ordered & 848 

\\
\hline
nice & 250 & best & 802 &  love & 198 & know & 829   \\
\hline
poor & 232 & few & 654 & served & 196 & make & 767  \\
\hline
awesome & 214 & friendly & 645 & visit & 160 & going & 743 \\

\end{tabular}}
\label{top_adjectives_verb}
\end{table}


\subsection{Performance of classical ML Classifiers using Linguistic and Lexical Features}
The linguistic analysis reveals that several attributes of reviews, such as review length, range of vocabulary used, coverage of opinion lexicon, and usage of part-of-speech (POS) vary across two demographic groups that can be attributed to the reviewer's English language nativeness or proficiency level and socio-culture (shown in Table \ref{content-independent-statistics}). It is observed that the usage of common opinion terms is more apparent (i.e., high coverage of sentiment lexicon)  in the writing of non-native speakers (Demography-1). Figure \ref{linguisti_classifiers} provides the precision, recall, F1 score, and accuracy of various classical ML-based methods utilizing linguistic features. We can see that when all the linguistic features (i.e., length, grammatical, and sentiment) are utilized, all of the four classical ML classifiers LR, SVM, RF, and ET perform similarly; They achieve F1 scores of around 0.833 and accuracy around 83\%.


We apply the Mann-Whitney test to find whether any of the stylistic features are significantly different in the reviews of \textit{Demography-1} and \textit{Demography-2}. A $p$-value of 0.05 is used for the significance test. The Mann-Whitney test indicates many of the linguistic features are significantly different in the two groups; however, we observe they do not provide very high distinguishing power when incorporated into the classical ML classifiers. For example, the text statistic features, such as review length and sentence length,  are significantly different ($p$-value less than 0.05) in both groups; however, when incorporated into the classical ML, they obtain a much lower F1 score of 0.813 than the lexical features. The high standard deviations (std.) of the review length, with respect to both word and sentence, indicate that many reviews in each group spread far away from its group mean value, which may induce the classifier to yield wrong predictions.

Classical ML classifiers yield F1 scores between 0.89 and 0.94 when word unigram and bigram lexical features are utilized (see Fig. \ref{lexical_cml}). The better performance of ML classifiers with lexical features indicates the presence of some distinguishing socio-culture-specific words and named entities in the reviews of both groups, which generate effective signals to discern the two groups. Nevertheless, we find that reviews of both groups share some common adjectives, such as \textit{good}, \textit{great}, \textit{nice}, and \textit{best} (Table \ref{top_adjectives_verb}), which are among the most frequently occurring adjectives in both groups. A few other adjectives are also common to both groups.

This observation suggests that both demographic groups, irrespective of English language nativeness, tend to use simple and commonly used adjectives to express opinions and feelings.
One dissimilarity we observe is that in reviews of \textit{Demography-2} (written mostly by native speakers), adjectives of quantity such as \textit{more}, \textit{little}, and \textit{few} are more frequent than in the \textit{Demography-1} review group. None of these adjectives of quantity appear among the top 10 adjectives in \textit{Demography-1}, and none of them occur more than 214 times.
When examining the verbs used in the two groups of reviews, we observe that the most frequent verbs are also very similar (see Table \ref{top_adjectives_verb}). Although there is a high presence of overlapping words in both groups, there are certainly distinguishing words, such as named entities, that help the classifiers discern between the two groups.





\subsection{Performance of Deep Learning based Classifiers}
The fine-tuned transformer-based language models yield impressive results utilizing unigram and bigram-based lexical features; they attain almost perfect accuracy by correctly classifying around 97\% instances (Fig \ref{lexical_all}). The pre-trained language models are generated based on an enormous amount of textual content, which helps to capture the implicit pattern of the reviews and can effectively identify the language nativeness of reviewers. Also, the high efficacy of lexical feature-based classical ML classifiers and transformer-based models can be attributed to the English language proficiency level of the people of Bangladesh, who are not known as very fluent English speakers. Additionally, the presence of named entities with specific meanings and socio-cultural terms in the reviews of both groups supports the lexical n-gram-based approach, leading to better performance.



\subsection{Implications of the Study}

This study provides insights into various perspectives by analyzing data from two demographic groups, including the followings-

\paragraph{Language Landscapes and Identity}
The study contributes to understanding how English language variation in social media usage is linked to different demographic groups. It can shed light on how individuals from diverse demographic backgrounds use language to express their cultural and social identities, providing insights into the complex relationship between language, ethnicity, and nationality.

\paragraph{Lexical Diversity in Online Communication} Examining the lexical diversity in online social media review data can shed light on the richness and variety of language used by individuals. This analysis can help researchers understand the level of vocabulary sophistication, linguistic creativity, or the influence of cultural factors in online communication.

\paragraph{Sociolinguistic Research and Language Policy}
The study's findings can contribute to sociolinguistic research and language policy discussions. Understanding the relationship between language nativeness, demographic diversity, and social media usage can inform discussions on language rights, language maintenance, and linguistic identity in digital spaces.

\section{Summary, Limitations and Future Work}
This study aims to distinguish reviews of two different socio-demographic groups leveraging various linguistic and lexical features, language models, and ML classifiers. From two demographically distinct review groups (\textit{Demography-1} and \textit{Demography-2}), various linguistic features are extracted to train ML classifiers for the demography prediction task. In addition, two state-of-the-art pre-trained language models, BERT and RoBERTa, are fine-tuned with the n-gram-based features for the prediction task. We observe that linguistic features are capable of distinguishing demographically diverse reviews; when they are fed into classical ML classifiers, an F1 score (best result) close to 0.85 is obtained. The pre-trained models exhibit very high efficacy for distinguishing reviews using lexical features, which can be attributed to the presence of name-entity and sociocultural-specific features in the two review groups. Our analysis reveals the contrast and similarity of implicit characteristics of reviews written by two demographically diverse review groups. We present multiple approaches for the demography prediction  task that can help a diverse set of downstream decision-making tasks.

One of the limitations of this work is that it only considers two demographically distinct groups. Since English proficiency levels (a predominant factor affecting linguistic characteristics) among non-native speakers may vary across demographics, such as geography, cultures, and language families, it is worthwhile to analyze non-native English review data from multiple demographics. Furthermore, it is worth exploring whether linguistic features exhibit similar distinguishing signals when demographics representing people with similar English language proficiency levels (e.g., native and highly proficient non-native speakers) are considered. In our future work, we will focus on collecting, annotating, and analyzing review data from diverse demographics, including variations in English fluency, geographical regions, native language families, and socio-cultures.

\bibliography{Reference}

\bibliographystyle{ieeetr}





\end{document}